\documentclass{article}
\usepackage{arxiv}
\usepackage{booktabs}
\usepackage{graphicx}
\usepackage{amsmath}
\usepackage{multirow}
\usepackage{algorithmicx}
\usepackage[linesnumbered, ruled, vlined]{algorithm2e}
\usepackage[titletoc, toc, page]{appendix}
\usepackage{natbib}
\usepackage[hidelinks]{hyperref}

\begin{document}

\title{Pattern Detection in the Activation Space for Identifying Synthesized Content}

\author{%
  Celia Cintas\thanks{The paper is under consideration at Pattern Recognition Letters}\\
  IBM Research\\
  Nairobi, Kenya\\
  \And
  Skyler Speakman\\
  IBM Research\\
  Nairobi, Kenya\\
  \And
  Girmaw Abebe Tadesse \\
  IBM Research\\
  Nairobi, Kenya\\
  \And
  Victor Akinwande\\
  IBM Research\\
  Nairobi, Kenya\\
  \And
  Edward McFowland III\\
  University of Minnesota\\
  Minneapolis, MN, USA\\
  \And
  Komminist Weldemariam\\
  IBM Research\\
  Nairobi, Kenya\\
}

\maketitle

\begin{abstract}


Generative Adversarial Networks (GANs) have recently achieved unprecedented success in photo-realistic image synthesis from low-dimensional random noise. The ability to synthesize high-quality content at a large scale brings potential risks as the generated samples may lead to misinformation that can create severe social, political, health, and business hazards.
We propose SubsetGAN to identify generated content 
by detecting a subset of anomalous node-activations in the inner layers of pre-trained neural networks. These nodes, as a group, maximize a non-parametric measure of divergence away from the expected distribution of activations created from real data. This enable us to identify synthesised images without prior knowledge of their distribution.
SubsetGAN efficiently scores subsets of nodes and returns the group of nodes within the pre-trained classifier that contributed to the maximum score. The classifier can be a general fake classifier trained over samples from multiple sources or the discriminator network from different GANs. Our approach shows consistently higher detection power than existing detection methods across several state-of-the-art GANs (PGGAN, StarGAN, and CycleGAN) and over different proportions of generated content. 
\end{abstract}


\section{Introduction}

The accelerated growth of deep learning models for synthetic generation, such as GANs~\citep{Goodfellow2014}, has made it possible to create near realistic fake content at a massive scale, generating thousands of samples in seconds. The generation capabilities range from full synthetic faces~\citep{karras2017progressive,karras2020analyzing}, or partial image modification, such as attribute editing~\citep{StarGAN2018,liu2019stgan} to image style transfer~\citep{CycleGAN2017}. The generated  samples from GANs were reported to be challenging for the human eye to distinguish from real samples\footnote{\url{https://thispersondoesnotexist.com/}}~\citep{karras2017progressive}. 
With near realistic generated content and high throughput capacity, several high-profile concerns are on the rise in critical areas such as security, ethics, democracy, and intellectual property rights.  If the trend continues, the traditional perspective of treating images (\textit{"A picture is worth a thousand words"}) as reliable and trustworthy content may no longer be valid. 
This will challenge data-driven decision making in societal and commercial activities.
Therefore, an effective technique to detect  fake (AI-synthesized) content is crucial for platforms that need to verify content from unknown sources.
Several methods for detecting counterfeit or synthetic content employ either ad-hoc forensics features or dedicated deep learning architectures to distinguish fake from real content in a given data type and form. These methods rely on labeled samples from different generative sources, data augmentation via replication processes, or specialized deep learning models and training techniques~\citep{hsu2020deep, zhang2019detecting}.
\begin{figure*}[t]
    \includegraphics[width=\textwidth]{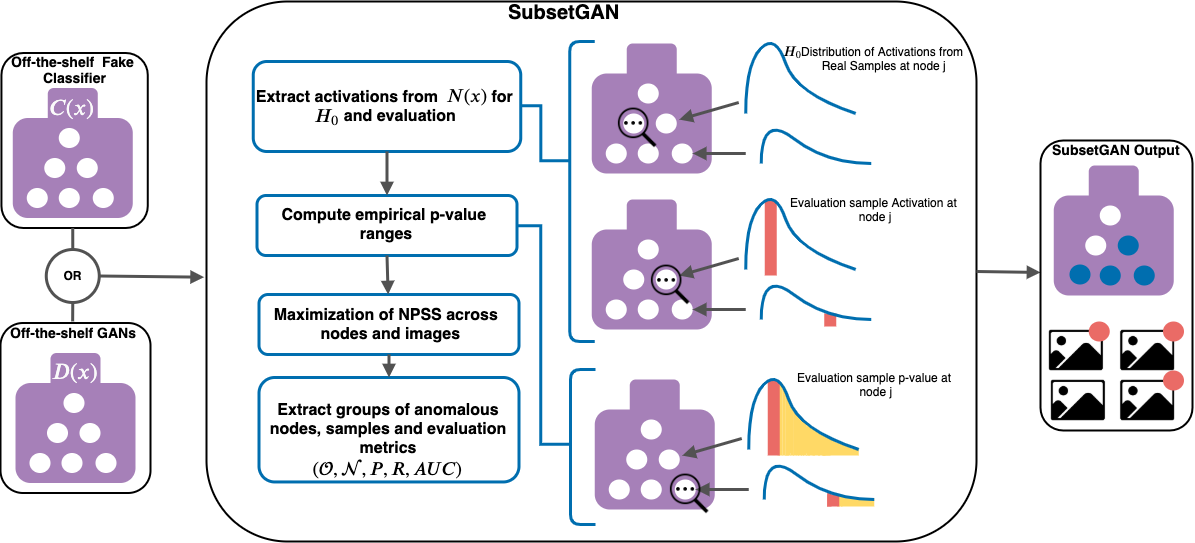}
    \caption{Overview of the proposed approach - SubsetGAN. First,  we analyze the distribution of the activation space of the given network. The model can be a pre-trained fake classifier $C(x)$ or the discriminator network of an off-the-shelf GAN $D(x)$. 
 After we extracted the activations from the model, we compute the empirical $p-$values followed by the maximization of non-parametric scan statistics - NPSS~\citep{feng-npss_graph-2014}. Finally, a subset of images and the corresponding anomalous subset of nodes in the network are identified.}
    \label{fig:approach}
\end{figure*}
In this paper, we propose SubsetGAN that determines whether a given batch of input samples contains a synthesized subset of samples using an anomalous pattern detection method called group-based subset scanning~\citep{neill-ltss-2012,mcfowland-fgss-2013}. This work builds on top of recent works that employ \emph{individual} scanning to detect adversarial attacks across audio and images~\citep{cintas2020detecting,akinwande2020identifying} without exploiting the patterns potentially shared across a group of samples. We hypothesize that synthesized content from off-the-shelf GANs (e.g., PGGAN \citep{karras2017progressive}, DCGAN \citep{radford2015unsupervised}, and StarGAN \citep{StarGAN2018}) leave a potentially subtle but systematic trace in the activation space across multiple generated samples. We test this hypothesis through group-based subset scanning over the activation space that encodes \emph{groups of samples} that may appear anomalous when analyzed together. In short, this work identifies which, of the exponentially-many, subset of samples in a test set have higher-than-expected activations at which, of the exponentially-many, subset of nodes in a hidden layer of a pre-trained neural network.
An overview of the proposed approach is shown in Figure~\ref{fig:approach}.

This work makes four main contributions. First, we show how to detect synthesised image content by applying group-based subset scanning methods on activations from internal layers of pre-trained models, including binary classifiers (trained to detect fake from real) and  state-of-the-art GANs.
Second, we present the unique ability to identify patterns of anomalous activations across a group of images.
Third, we validated our detection capabilities for both total and partial generation of images  from one to multiple generation sources. 
Fourth, we enhance the performance of the discriminant component of GANs and off-the-shelf fake classifiers to detect the synthesized images without extra labeled examples, data augmentation, or model retraining. 
We compare the proposed  SubsetGAN
 with state-of-the-art methods, such as FakeSpotter \citep{wangfakespotter} and AutoGAN \citep{zhang2019detecting}. SubsetGAN is also validated across the state-of-the-art generative models (see Tables~\ref{tab:sota}, \ref{table:innersubsetStarGAN} and \ref{tab:precisionpggan}). We 
show that the proposed group-based scanning method achieves higher synthesized content detection power compared to existing methods, tested under different types of content generation such as attribute editing, image translation, and full synthetic samples.


\section{Related Work}\label{sec:gans}

Generation of adversarial content can be categorized into two groups: \textit{complete (full)} and \textit{partial}. Complete generation is the case where a GAN generates the output from a noise vector~\citep{emami2018generating,han2018gan}. On the other hand, partial generation is a case where only a part of the content is generated or modified. Examples of partial generation include \textit{image translation}, and \textit{attribute editing}. Image translation refers to taking images from one domain and transforming them to have the style of images from another domain. 
Lastly, attribute editing refers to a case where only the input sample is modified in a specific way for a given characteristic. 
Next, we review existing methods related to adversarial content generation and the capability of recent techniques to detect them.

\subsection{Generative Adversarials Networks (GANs)}
GANs have shown remarkable results in various computer vision tasks such as image generation~\citep{emami2018generating,han2018gan}, image translation~\citep{zhu2017toward,armanious2020medgan}, face image synthesis~\citep{karras2017progressive,ye2019triple,zhang2018stackgan++} and recently in generation of text ~\citep{chen2018adversarial,xu2018diversity} and audio~\citep{lorenzo2018can}.
A typical GAN framework contains generative ($G$) and  discriminative ($D$) components such that $G$ aims to generate realistic-like content while $D$  learns to discriminate if a generated sample is from the real data distribution ($H_0$) or not ($H_1$). Multiple iterations will inform  $ G $ on how to adjust the generation process to fool $ D $ and vice-versa.

We provide a summarised review of the existing methods related to the generation of adversarial content used in this work: DCGAN~\citep{radford2015unsupervised}, PGGAN~\citep{karras2017progressive},  StarGAN~\citep{StarGAN2018}, and  CycleGAN~~\citep{CycleGAN2017}. 
Deep Convolutional Generative Adversarial Networks (DCGAN)~\citep{radford2015unsupervised} is one of the popular and successful networks designed for GAN and commonly used as a standard for  image generation . It mainly composes of convolution layers without max pooling or fully connected layers. 
 
More recently, \citep{karras2017progressive} presented PGGAN - a new direction for full image synthesis that enhanced the quality of generated images by employing a progressive training approach that fine-tunes layers increasingly over time, and a simple technique to increase variation across the generated images without learnable parameters. Furthermore, PGGAN also provided a novel evaluation technique that takes into consideration both the quality of generated images and their variations. 

StarGAN~\citep{StarGAN2018} is a multiple attribute editing network and its design is aimed at improving the scalability and robustness via a novel GAN that learns the mappings among multiple domains. StarGAN uses only single generator and single discriminator to effectively train a model from images of multiple domains.  

Lastly, CycleGAN~\citep{CycleGAN2017} is an image translation network that provided a means for image translation in the absence of paired training data, by exploiting the cycle consistent translation property between source and target domains and later combining the disparity from this property with the standard adversarial loss. CycleGAN was validated across different applications, such as collection style transfer, object transfiguration, season transfer and photo enhancement. 



\subsection{AI-synthesized Content Detectors}
Existing methods for detecting synthesized content might use forensic-based approaches or deep learning techniques. Several works from these two streams are analyzed below.  Raw pixels and ad-hoc forensics features, extracted from real and generated content, were used to train a classifier in \citep{marra2018detection}. In \citep{hsu2018learning}, the authors proposed a contrastive loss in seeking the typical features of the synthesized images followed by  a classifier to detect such AI-generated images. Similarly, \citep{hsu2020deep} reported a deep learning-based approach for identifying the fake images by using the contrastive loss. A simulator was employed to generate images and a spectrum-based classifier was used in \citep{zhang2019detecting}. 
Closer to our work, FakeSpotter~\citep{wangfakespotter} was designed to monitor node behavior to detect generated faces since patterns of layer-by-layer node activation  may capture more subtle features that are crucial to detect synthesized content, using a threshold calculated from the average values of outputs in each layer for a given set of training samples. 

We hypothesize that more complex layer-by-layer activation behavior underlies learned representations in deep neural networks. Thus, we can enhance average threshold information with a subset of nodes in each layer that yield higher-than-expected activations under generated content. 
SubsetGAN provides a way to simultaneously detect and characterize multiple samples created by GANs. 
Unlike the state-of-the-art adversarial content detection techniques, our proposed approach does not require extra labeled examples, data simulation processes, or specialized training techniques, which must be asserted before training time. These capabilities of the proposed approach could also be employed to  enhance the performance of existing deep learning methods. 

\section{Proposed Approach: SubsetGAN}\label{sec:groupscan}

Subset scanning treats the 
detection problem as a search for the {\em most anomalous} subset of observations in the data.  This exponentially large search space is efficiently explored by exploiting mathematical properties of our measure of anomalousness.

Consider a set of images  $X =\{X_1 \cdots X_M\}$ and nodes $O = \{O_1 \cdots O_J\}$ within the discriminator $D$ or pre-trained classifier $C$.  Let $X_S \subseteq X$ and $O_S \subseteq O$,  we then define the subsets $S$ under consideration to be
$S = X_S\times O_S$. The goal is to find the most anomalous subset:

\begin{equation}
    S^{*}=\arg \max _{S} F(S)
\end{equation}
where the score function $F(S)$ defines the anomalousness of a subset of images and node activations.   
SubsetGAN uses an iterative ascent procedure that alternates between two steps: a step identifying the most anomalous subset of images ($X_s$) for a fixed subset of nodes ($O_s$), or a step that identifies the converse.
There are $2^M$ possible subsets of images, $X_S$, to consider at these steps.  However, the Linear-time Subset Scanning property (LTSS) \citep{neill-ltss-2012,speakman_penalized} reduces this space to only $M$ possible subsets while still guaranteeing that the highest scoring subset will be identified.  This drastic reduction in the search space is the key feature that enables SubsetGAN to scale to large networks and sets of images.  Without loss of generality, LTSS also decreases the search space of node subsets from $2^J$ to $J$ at each of the remaining steps of the ascent procedure.  The iterations will converge to a joint, local maximum such that any change to the subset $X_S$, \emph{conditioned} on the subset $O_S$, decreases the score $F(S)$.  Similarly, any changes to $O_S$ conditioned on $X_S$ will also decrease $F(S)$.  Multiple random restarts are used to approach a global maximum.

\subsection{Non-parametric Scan Statistics (NPSS)}

SubsetGAN uses 
NPSS that has been used in other pattern detection methods \citep{mcfowland-fgss-2013,mcfowland-tess-2018,feng-npss_graph-2014}.  Given that NPSS makes minimal assumptions on the underlying distribution of node activations, SubsetGAN has the ability to scan across different type of layers and activation functions. 
However, these methods do require baseline or background data to inform their data distribution under the null hypothesis $H_0$ of no generated content present.

\begin{figure*}[t]
    \centering
    \includegraphics[width=0.9\textwidth]{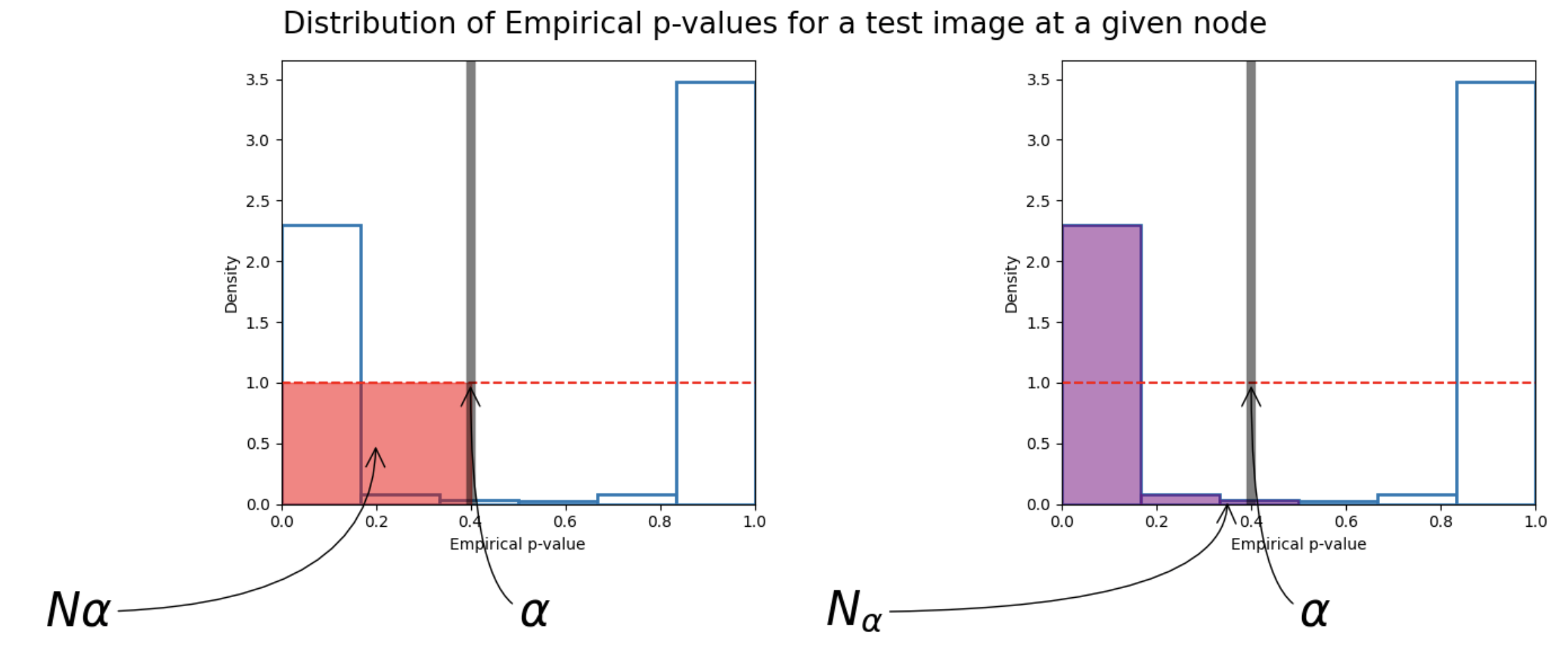}
    \caption{Example of an evaluation image at a given node in order to measure how much the p-values deviate from uniform. Where $N_\alpha$ is the number of p-values less than $\alpha$, $N$ is the number of p-values, and $\alpha$ is the level of significance.}
    \label{fig:distpvalues}
\end{figure*}

There are three steps to use non-parametric scan statistics on model's activation data. The first is to form a distribution of  ``expected''  activations at each node.  We generate the distribution by letting the discriminator process samples that are known to be real (sometimes referred to as ``background'' samples) and record the activations at each node.  The second step 
records the activations induced by the group of test images and compares them to the baseline activations created in the first step.  This comparison results in a $p$-value at each node, for each image in the test set (Eq.~\ref{eq:p_vals}). 
See Figure ~\ref{fig:distpvalues} for an example of this process. Lastly, we quantify the anomalousness of the resulting $p$-values by finding $X_S$ and $O_S$ that maximize the NPSS, which quantify  how much an observed distribution of $p$-values deviates from the uniform distribution. 
A visual overview of these three steps is shown in Figure~\ref{fig:approach}.
 

Let $A^{H_0}_{zj}$ be the matrix of activations from $Z$ real images at each of $J$ nodes in a discriminator layer.  Let $A_{ij}$ be the matrix of activations induced by $M$ images in the test set, that may or may not be generated.  
SubsetGan computes an empirical $p$-value for each $A_{ij}$, as a measurement for how anomalous the activation value of a potentially generated image $X_i$ is at node $O_j$. 
This $p$-value $p_{ij}$ is the proportion of activations from the $Z$ background images, $A^{H_0}_{zj}$, that are larger or equal to the activation from an evaluation image 
 at node $O_j$.
\begin{equation}
    p_{ij} = \frac{1+\sum_{z=1}^{|Z|} I(A^{H_0}_{zj} \geq A_{ij} )}{|Z|+1}\label{eq:p_vals}
\end{equation}
Where $I(\cdot)$ is the indicator function. A shift is added to the numerator and denominator so that a test activation that is larger than \emph{all} activations from the background at that node is given a non-zero $p$-value.  Any test activation smaller than or tied with the smallest background acivation at that node is given a $p$-value of 1.0. 







SubsetGAN processes the matrix of  $p$-values ($P$) from test images with a NPSS to identify  a submatrix $S =X_S \times O_S$ that maximizes  $F(S)$, as this is the subset with the most statistical evidence for having been affected by an anomalous pattern. 
The general form of the NPSS score function is 
\begin{equation}
F(S)=\max_{\alpha}F_{\alpha}(S)=\max_{\alpha}\phi(\alpha,N_{\alpha}(S),N(S))
\end{equation}
where $N(S)$ is the number of empirical $p$-values contained in subset $S$ and $N_{\alpha}(S)$ is the number of $p$-values less than (significance  level) $\alpha$ contained in subset $S$. 
It has been shown that for a subset $S$ consisting of $N(S)$ empirical $p$-values, $E\left[N_{\alpha}(S)\right] = N(S)\alpha$ ~\citep{mcfowland-fgss-2013}.
SubsetGAN attempts to 
find the subset $S$ that shows the most evidence of an observed significance higher than an expected significance, 
$ N_{\alpha}(S) > N(S)\alpha $, for some significance level $\alpha$.

We compare the use of two statistical tests for generated content detection: Berk-Jones (BJ) and Higher-Criticism (HC). BJ test statistic~\citep{berk-bj-1979} is defined as:
\begin{equation}
    \phi_{BJ}(\alpha,N_\alpha,N) = N*{KL} \left(\frac{N_\alpha}{N},\alpha\right)
\end{equation}
where $KL$ refers to the Kullback-Liebler divergence, $KL(x,y) = x \log \frac{x}{y} + (1-x) \log \frac{1-x}{1-y}$, between the observed and expected proportions of significant $p$-values. We can interpret BJ as the log-likelihood ratio for testing whether the $p$-values are uniformly distributed on $[0,1]$.

The second statistic, Higher-Criticism~\citep{donoho2004higher} is defined as 
\begin{equation}
\phi_{HC}(\alpha, N_\alpha, N) = \frac{|N_\alpha - \alpha N|} {\sqrt{N\alpha(1-\alpha)}}.    
\end{equation}
This can be interpreted as the test statistic of a Wald test for the amount of significant $p$-values given that  $N_{\alpha}$ is binomially distributed with parameters $N$ and $\alpha$ under $H_0$. 
Because HC normalizes by the standard deviation of $N_{\alpha}$, it tends to return small subsets with extreme $p$-values. 
\subsection{Efficient maximization of NPSS}
\begin{algorithm}[t]
\caption{Single Restart over $M$ test images and $J$ nodes }\label{alg:singlerestart}
\SetKwFunction{OptimizeRows}{OptimizeRows}
\SetKwFunction{Random}{Random}
\SetKwInOut{Input}{input}
\SetKwInOut{Output}{output}

\Input{$(M \times J)$ $p$-values}
\Output{score, $X_s$, $O_s$}
score $\leftarrow -1$ \;
$X_s \leftarrow$ \Random{$M$} \;
$O_s \leftarrow$  \Random{$J$} \;
\While{score is increasing }{
$(M \times J') = (M \times J) | O_s$ \; 
score, $X_s \leftarrow$  \OptimizeRows{($M \times J'$)} \; 
$(M' \times J) = (M \times J) | X_s$ \; 
score, $O_s \leftarrow$  \OptimizeRows{$(J \times M')$}\; 
} 
\Return{score, $X_s$, $O_s$}
\end{algorithm}
\begin{algorithm}[h!]
\caption{Optimize over rows using LTSS ({\em OptimizeRows}). It maintains maxscore and $arg\_max\_subset$ over $\|E\|*\|T\|$ subsets.} \label{alg:optimize_rows}
\SetKwFunction{SortByPropCT}{SortByPropCT}
\SetKwFunction{LinearSpace}{LinearSpace}
\SetKwInOut{Input}{input}
\SetKwInOut{Output}{output}
\Input{$p$-values from all rows E and relevant cols C}
\Output{maxscore, $arg\_max\_subset$}
maxscore $\leftarrow -1$\; $arg\_max\_subset \leftarrow \emptyset$ \;
\For{$t$  in T = \LinearSpace{0,1}}{ 
$sorted\_priority \gets$ \SortByPropCT{E, t}  \Comment{/* Sort elements in $E$ by proportion of $p$-values across $C < t$}.   */ 

 $Score(sorted\_priority, t)$  \Comment{/* Score $|E|$ subsets of $sorted\_priority$ by iteratively including elements one at a time. */} \;
}
\Return maxscore, $arg\_max\_subset$
\end{algorithm}

SubsetGAN identifies the anomalous subset of $p$-values through iterative ascent of two optimization steps, see Algorithm \ref{alg:singlerestart}. Within each step, the number of subsets to consider is reduced from $O(2^E)$ to $O(E)$ where $E$ is the number of elements currently being optimized, either images or nodes, see Algorithm~\ref{alg:optimize_rows}.  This efficient optimization is a direct application of the LTSS property \citep{neill-ltss-2012,speakman_penalized}.  Each element $e$ is sorted by its priority, which is its proportion of $p$-values less than an $\alpha$ threshold.  Once sorted, the LTSS property states that the highest-scoring subset will consist of the top-$k$ elements for some $k$ between 1 and $|E|$. Any subset not consisting of the top-$k$ priority elements is sub-optimal and therefore does not need to be evaluated.

\begin{small}
\hspace{-0.2cm}

\end{small}
\section{Experimental Setup}
We evaluated the performance of SubsetGAN using multiple score functions, under different experimental scenarios and compared against the state-of-the-art synthesized-content detection methods. The scenarios include different generation types, i.e., complete and partial. 
Individual-input scanning was used as our baseline. We experimented with the proposed SubsetGAN on the activations drawn from the discriminator components of different pre-trained GANs. SubsetGAN was also validated SubsetGAN on pre-trained classifiers that were trained with samples from multiple GANs 
(e.g., 
fake face detectors based on ResNet or SqueezeNet). This helps to provide a scenario that does not require knowledge of the source of the generated content, as is the case for scanning over the discriminator. We also visualized the set of nodes in the discriminators that behaved differently for the anomalous groups. 
The metrics used are area under receiver operating characteristic curve (AUC), precision ($P$) and recall ($R$). 
In group-scanning results, AUC can be thought of as detection \emph{power}, which is the method's ability to distinguish between test sets that contain some proportion of synthetic images and test sets containing only real content.  P and R reflect detection performance, which is the method's ability to label which images in the test set are synthetic.   



\subsection{Datasets, GANs and Methods for Comparison}
For our experiments, we used images  from CelebA HQ \citep{liu2015faceattributes}, MNIST ~\citep{lecun1998gradient} and CycleGAN datasets ~\citep{CycleGAN2017} as the background samples.
To ensure high-quality and variety of generated samples, we used the pre-trained, highly cited GANs (Section~\ref{sec:gans}) for each of the generation types considered. 

To generate completely synthesized samples, we used a pre-trained PGGAN~\citep{karras2017progressive} model. We selected attribute editing and image translation as  use cases for partial generation of samples, and to this end, we used StarGAN~\citep{StarGAN2018} with two different sets of attributes. The first set contains five attributes (\textit{Black Hair, Blond Hair, Brown Hair, Male, and Young}). The second set contains only the \textit{Brown Hair} attribute as proposed by \citep{wangfakespotter}, for a fair comparison with other methods. For image translation, we used \textit{horse2zebra} model and its dataset as defined in~\citep{CycleGAN2017}.
Moreover, we also evaluated SubsetGAN over classic DCGAN using benchmarking datasets, such as MNIST. 

SubsetGAN scanned over the activations extracted from the discriminator components of the following pre-trained GAN models: DCGAN, CycleGAN\footnote{\url{https://modelzoo.co/model/pytorch-cyclegan-and-pix2pix}}, PGGAN\footnote{\url{https://pytorch.org/hub/facebookresearch\_pytorch-gan-zoo\_pgan/}} and StarGAN\footnote{ \url{https://modelzoo.co/model/stargan}}. 
Furthermore, to present a more general scenario similar to ~\citep{wangfakespotter}, we scanned over the activations from two available universal pre-trained fake classifier based on Resnet18~\citep{he2016deep} and SqueezeNet~\citep{iandola2016squeezenet} architecture and weights. We retrained the model with partial and complete generated samples with accuracy over test samples of $0.912$ and $0.920$ respectively.
In order to compare the synthesized-content detection power of SubsetGAN, we used recently published works:  FakeSpotter~\citep{wangfakespotter}, specially designed for fake face detection, and AutoGAN~ \citep{zhang2019detecting}, designed for the detection of generic image manipulations.

\subsection{Subset scanning setup}
We run individual and group-based scanning on node activations extracted from several types of layers, which include PGGAN's Scale, GroupScaleZero and Decision layers in addition to common layers such as Conv2D and BatchNorm2D.
We run group-based scanning across several proportions of generated content in a group, ranging from $10\%$ to $50\%$.
 We used $Z=5000$ images to obtain the background activation distribution ($A^{H_0}$) for experiments regarding PGGAN, StarGAN and DCGAN,. For evaluation, each test set had images drawn from a larger set of $1000$ real images (separate from Z) and from 1500 generated samples.  The proportion represented in a test set varied in our experiments.  
For CycleGAN, we followed the same proportion with the sample size from the \emph{horse2zebra} dataset~\citep{CycleGAN2017}.
\begin{table}[t]
\centering
 \caption{{\bf Detection Power} (AUC) for group-based and individual subset scanning with two different score functions: BJ and HC, experimented over all layers of DCGAN's discriminator, under two different proportions for MNIST dataset.}
\begin{tabular}{lccc|ccc}
\toprule
\multicolumn{1}{c}{Layers} &
      \multicolumn{6}{c}{group-based and indiv. Subset scan}\\
      \midrule
      &
      \multicolumn{3}{c|}{Score Func. BJ} &
      \multicolumn{3}{c}{Score Func. HC} \\
      & \multicolumn{1}{c}{50\%} &
       \multicolumn{1}{c}{30\%} &
      \multicolumn{1}{c|}{Indv.} &
 \multicolumn{1}{c}{50\%} &
      \multicolumn{1}{c}{30\%} &
      \multicolumn{1}{c}{Indv.} \\
 \midrule
Conv2d\_1 & $1.0$    & $1.0$             & $1.0$   & $1.0$   & $1.0$    & $1.0$   \\
LReLU\_1  & $1.0$    & $1.0$             & $1.0$   & $1.0$   & $1.0$    & $1.0$   \\
Conv2d\_2 & $1.0$    & $0.781$  & $0.956$ & $1.0$   & $ 0.758$ & $0.661$ \\
BN2d\_1   & $1.0$    & $0.803$  & $0.956$ & $1.0$   & $0.778$  & $0.661$ \\
LReLU\_2  & $1.0$    & $ 0.795$ & $0.956$ & $1.0$   & $0.781$  & $0.661$ \\
Conv2d\_3 & $ 0.378$ & $0.383$           & $0.528$ & $0.487$ & $0.531$  & $0.523$ \\
BN2d\_2   & $0.408$  & $0.414$           & $0.528$ & $0.545$ & $0.508$  & $0.523$ \\
LReLU\_3  & $ 0.380$ & $0.425$           & $0.528$ & $0.503$ & $0.455$  & $0.523$ \\
Conv2d\_4 & $0.062$  & $0.191$           & $0.380$ & $0.284$ & $0.422$  & $0.380$ \\
Sigmoid   & $0.059$  & $0.185$           & $0.380$ & $0.298$ & $0.373$  & $0.380$ \\
\bottomrule
\end{tabular}
\label{table:innersubsetMNIST}
\end{table}
\begin{figure}[t]
    \centering
    \includegraphics[width=0.46\textwidth]{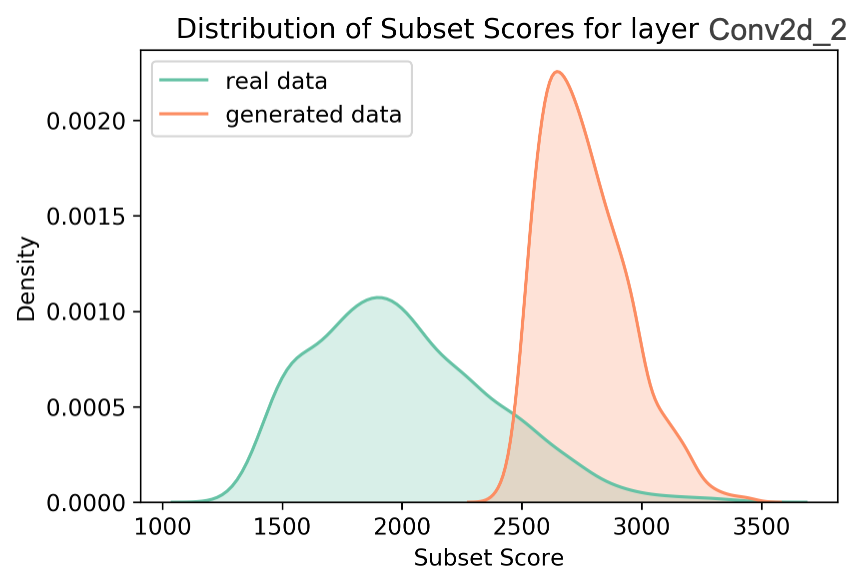}
    \includegraphics[width=0.47\textwidth]{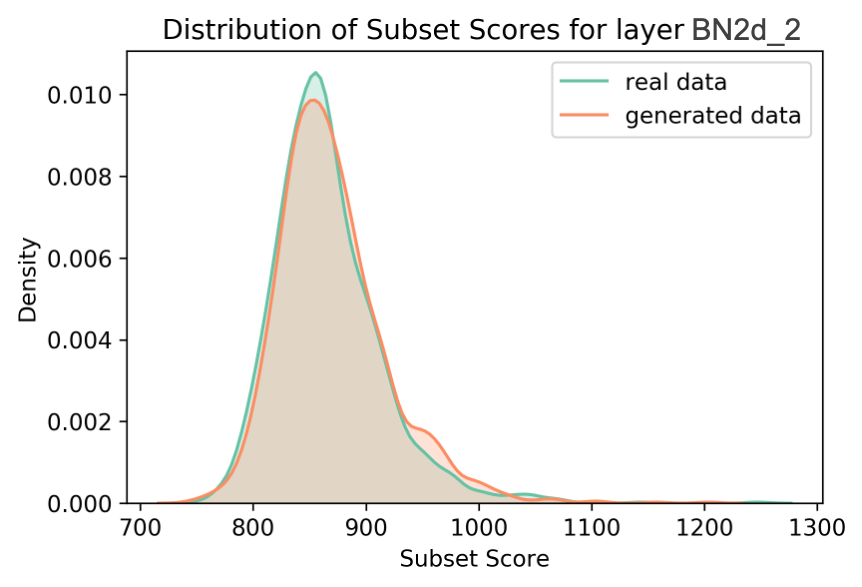}
    \caption{
    Subset scores distribution across layers of DCGAN $D(x)$ with BJ score function for real and generated samples. The distributions of individual subset scanning scores are shown in green for real images (expected distribution), and in orange for generated samples. Higher AUCs are expected when distributions are separated from each other (See layer Conv2d\_2) and lower AUCs when they overlap (See BN2d\_2). The computed AUC for the subset score distributions can be found in Table~\ref{table:innersubsetMNIST}.}
    \label{fig:subsetdist}
\end{figure}
\begin{figure}[h!]
    \centering
    \includegraphics[width=0.5\textwidth]{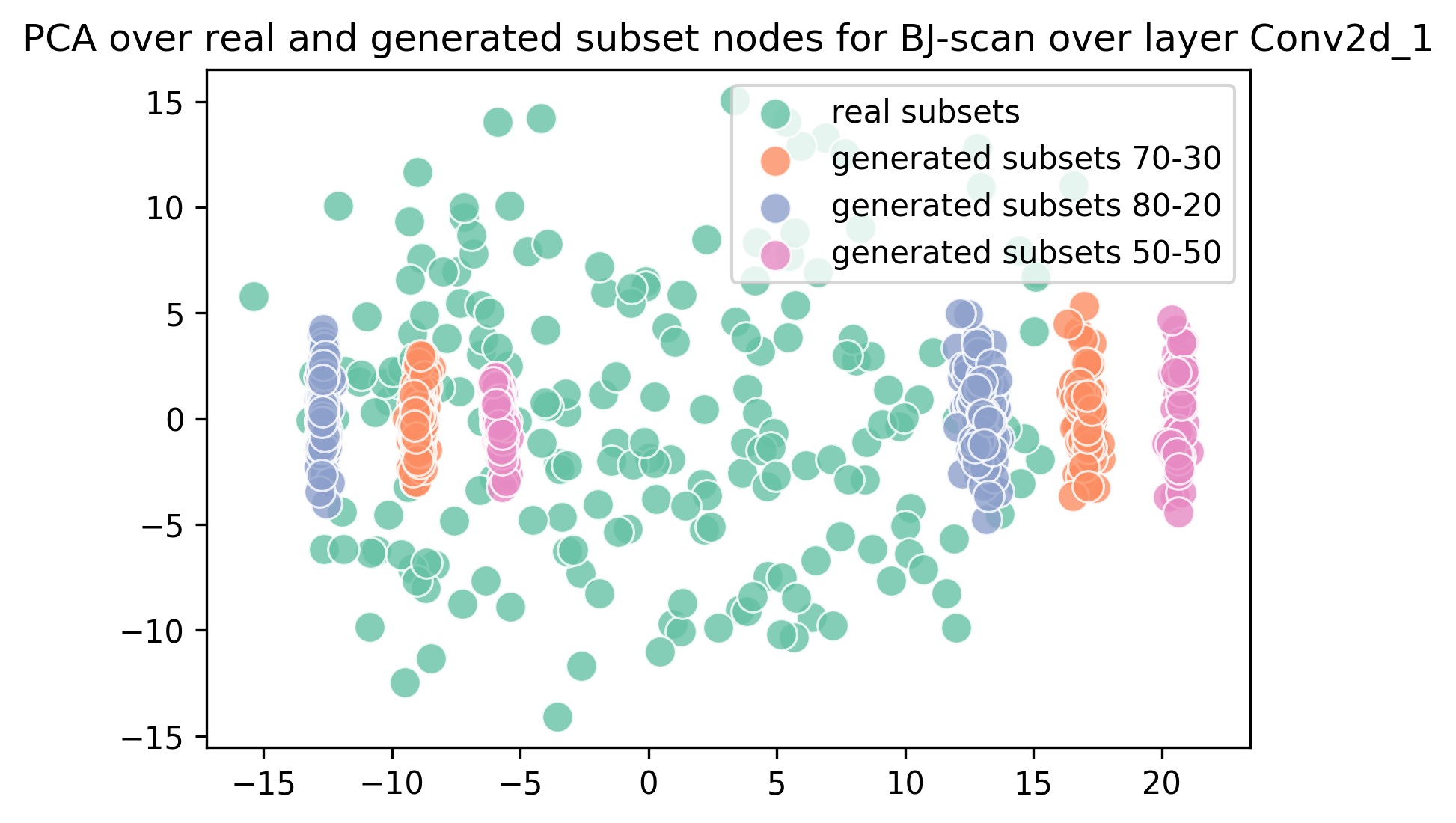}\includegraphics[width=0.5\textwidth]{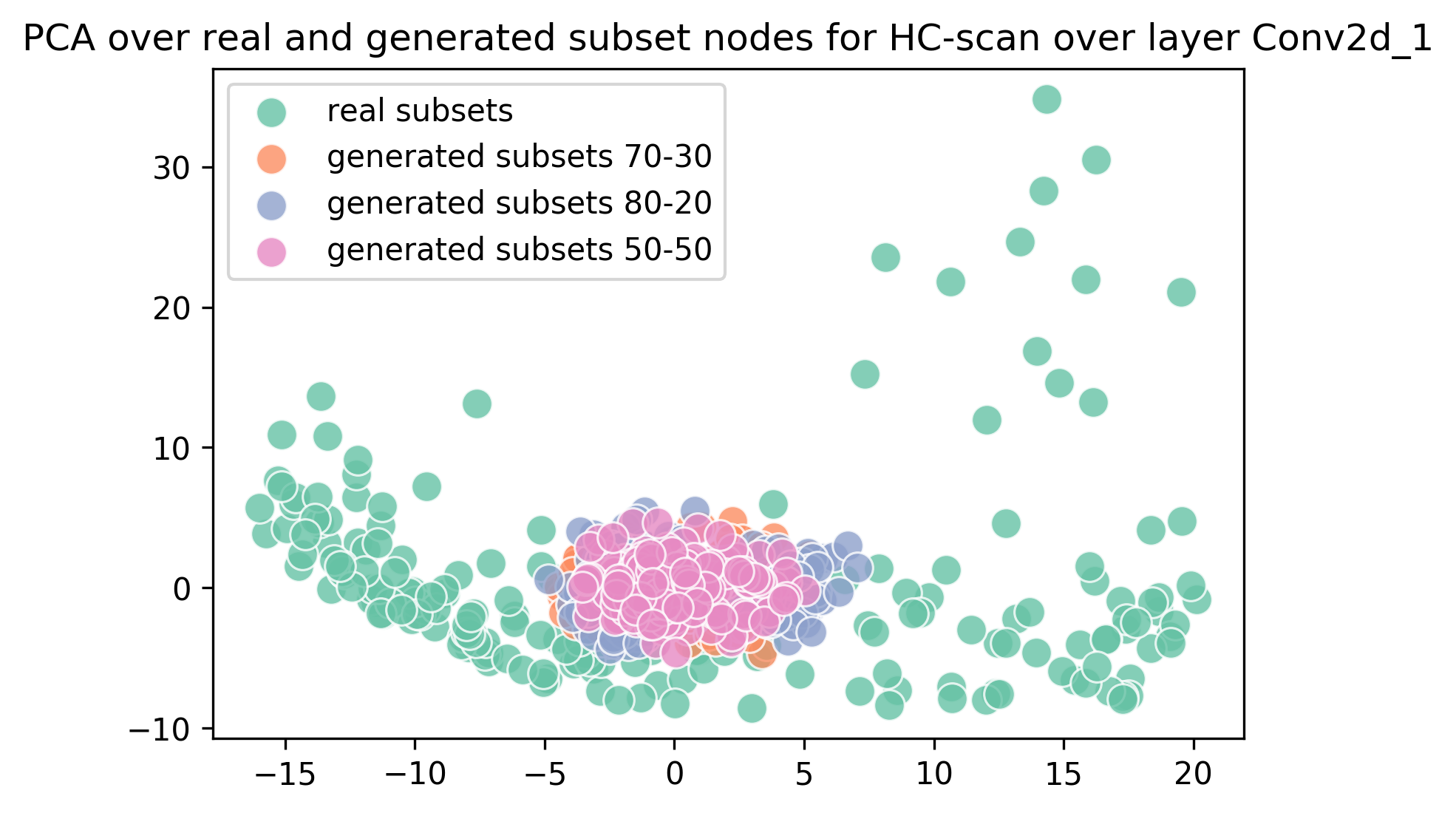}
    \caption{PCA for subsets of nodes for two different score functions 
    BJ and 
    HC from DCGAN over MNIST dataset. In Conv2d\_1, we can observe distinctive anomalous subsets of nodes, as we have perfect AUCs at that layer (See Table~\ref{table:innersubsetMNIST}). 
}
\label{fig:hcbjnodes} 
\end{figure}
\begin{table*}[t!]
\caption{Comparison, in \textbf{Precision (P) and Recall (R)},  of the proposed and existing adversarial detection methods under different Generation Types (GT): TS: total synthesis, AE: attribute editing.} 
\centering
\begin{tabular}{ccc|cc|cc}
\toprule
 
&  \multicolumn{6}{c}{Detection methods} \\

 &
      \multicolumn{2}{c}{SubsetGAN}& \multicolumn{2}{c}{FakeSpotter~\citep{wangfakespotter}} & \multicolumn{2}{c}{AutoGAN~\citep{zhang2019detecting}}\\
      \midrule
      \multicolumn{1}{p{2.cm}}{GT} 
      & {P} & {R} 
      & {P} & {R} 
      & {P} & {R} 
      \\
      \midrule
       TS (PGGAN) 
       & 0.941& 0.900 
       & \textbf{0.986} & \textbf{0.987} 
       & 0.926 & 0.974
       \\ 
       AE (StarGAN) 
       & \textbf{0.998} & \textbf{0.999} 
       & 0.901 & 0.865
       & 0.690& 0.567 
       \\ 

\bottomrule
\end{tabular}
\label{tab:sota}
\end{table*}

\begin{table}[]
 \caption{Detection Power for various detection methods with multiple network settings for different generation types: TS: total synthesis, AE: attribute editting.}
    \centering

    \begin{tabular}{cccc}
    \toprule
        Method & Generation Type & Network &  AUC \\
        \midrule
         FakeSpotter & TS & Fake face classifier & 0.985 \\
         FakeSpotter & AE & Fake face classifier & 0.881 \\ 
         \midrule
         AutoGAN & TS & GAN & 0.948 \\
         AutoGAN & AE & GAN & 0.656 \\
         \midrule
         SubsetGAN (indv) & TS & $D(x)$ from PGGAN & 0.950 \\
         SubsetGAN (indv) & AE & $D(x)$ from StarGAN & 0.999 \\
         \midrule
         SubsetGAN (group) &  TS & $D(x)$ from PGGAN & 0.999 \\
         SubsetGAN (group) &  AE & $D(x)$ from StarGAN & 1. \\
         SubsetGAN (group)&  AE \& TS & Fake classifier (ResNet) & 0.941 \\
         SubsetGAN (group) &  AE \& TS & Fake  classifier (SqueezeNet) & 0.994 \\
       
      \bottomrule   
    \end{tabular}
   
    \label{tab:generalclass}
\end{table}
\section{Results}

We compared detection power of two different non-parametric measures of anomalousness: Berk-Jones (BJ) and Higher Criticism (HC) (see Section~\ref{sec:groupscan}) for MNIST data generated by DCGAN.  
Table~\ref{table:innersubsetMNIST} shows AUC for both measures and two  proportions (50\% and 30\%) of synthetic content in each test set.  
Berk-Jones provides better detection power particularly when scanning over individual images only.  Scanning over earlier layers in the discriminator provides better detection power than deeper layers. 
An example of subset scores distributions for this network can be seen in Fig~\ref{fig:subsetdist}.

To better understand the subsets of anomalous nodes that were identified by SubsetGAN, we used dimensionality reduction techniques to visualise which subsets were similar-to or distinct-from each other.  We performed principal component analysis on the vector representations of the subsets of nodes.   
Figure~\ref{fig:hcbjnodes}, shows each subset of nodes returned by SubsetGAN across the top-2 principal components.  We observe that HC returns one consistent group of nodes when synthetic inputs are in the test set 
for both proportions while in BJ varies across ratios but with similar patterns. Furthermore, we observe the randomness of the anomalous nodes when the test set contains all real images. 
Under this condition, SubsetGAN does not identify any consistent group of nodes with higher activations. 
The rest of our experiments in this paper are all executed with the BJ score function to measure anomalousness.
Below, we provide the results achieved by SubsetGAN and existing methods in detecting \textit{partially}, and \textit{completely} synthesized samples. 
We have also evaluated the proposed SubsetGAN across different generation types, discriminator networks and universal fake classifiers as shown in Table~\ref{tab:generalclass}.

\begin{figure}[t!]
    \centering
   \includegraphics[width=0.46\textwidth]{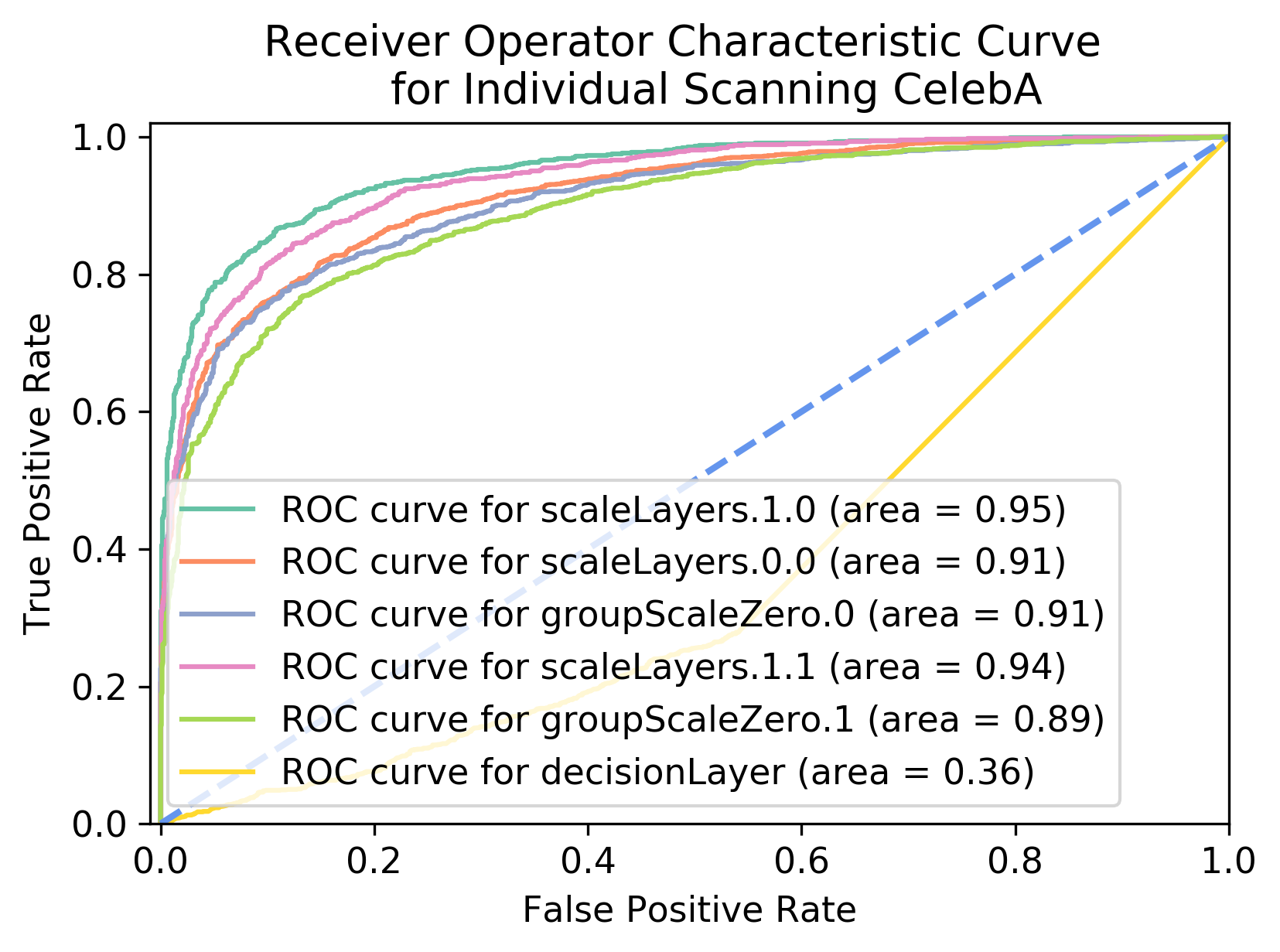}\includegraphics[width=0.46\textwidth]{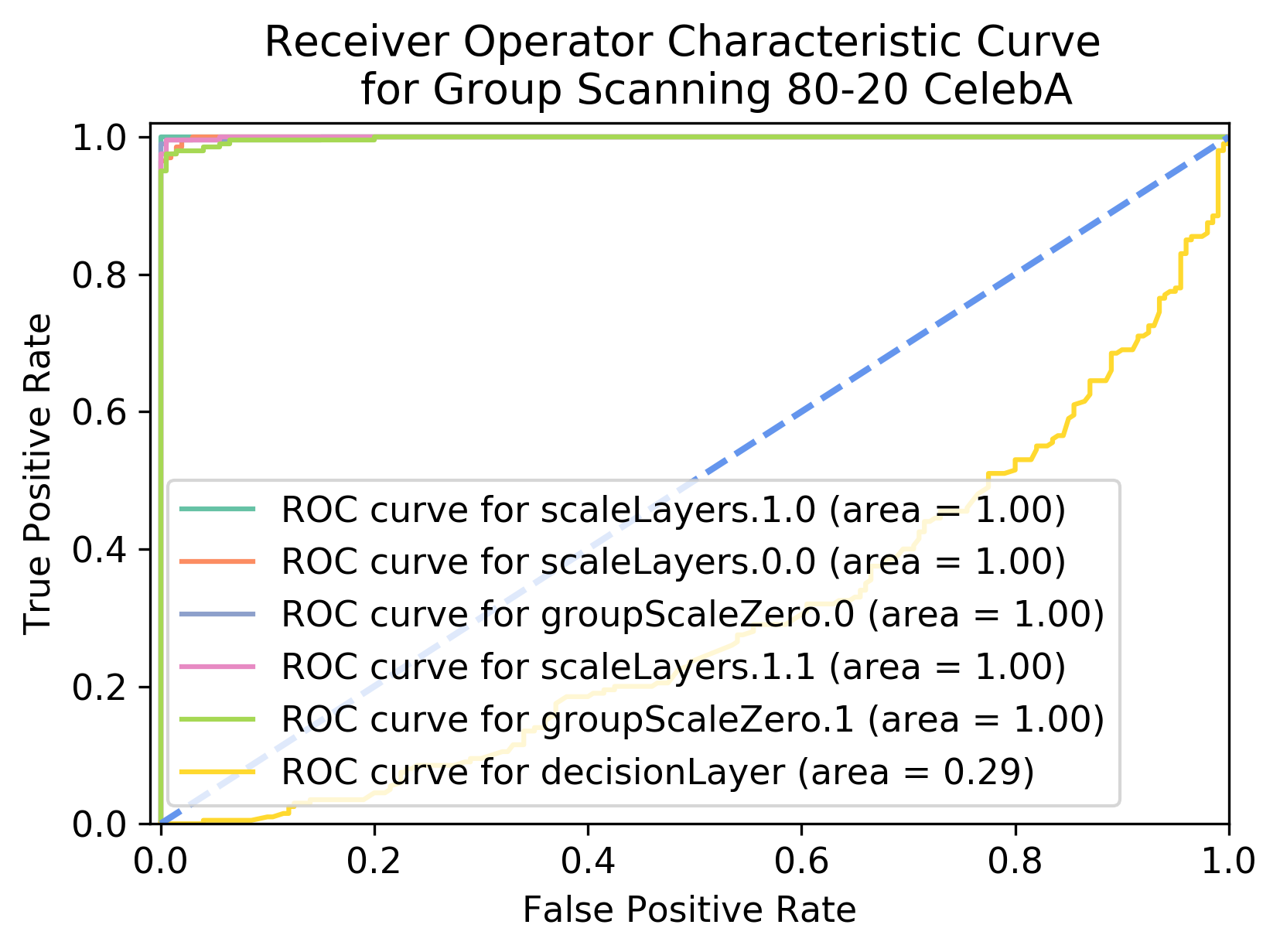} 
    \caption{ROC curves for individual and group scanning over 80\%-20\% proportion for PGGAN with CelebA-HQ.}
    \label{fig:rocpgganceleba}
\end{figure}
\begin{table}[t!]
\centering
\caption{{\bf Detection power} (in AUC) for group-based and individual subset scanning over all layers of StarGAN $D(x)$ under two different proportions (50\% and 20\%) for  detection of edited attributes that include Black Hair, Blond Hair, Brown Hair, Male and Young \citep{StarGAN2018}.}
\begin{tabular}{@{}lccc@{}}
\toprule
 &
      \multicolumn{3}{c}{Scanning} \\
      \cline{2-4}
      &\multicolumn{2}{c}{Group} & \multicolumn{1}{c}{}\\
        \multicolumn{1}{c}{Layers}& 50\%    & 20\%    &  Indv.   \\ \midrule
Conv2d\_1    & $1.0$   & $0.478$ & $0.225$ \\
LeakyReLU\_1 & $1.0$   & $0.694$ & $0.225$ \\
Conv2d\_2    & $1.0$   & $0.997$ & $0.351$ \\
LeakyReLU\_2 & $1.0$   & $0.995$ & $0.351$ \\
Conv2d\_3    & $1.0$   & $1.0$   & $0.961$ \\
LeakyReLU\_3 & $1.0$   & $1.0$   & $0.961$ \\
Conv2d\_4    & $1.0$   & $1.0$   & $0.992$ \\
LeakyReLU\_4 & $1.0$   & $1.0$   & $0.992$ \\
Conv2d\_5    & $1.0$   & $1.0$   & $0.997$ \\
LeakyReLU\_5 & $1.0$   & $1.0$   & $0.997$ \\
Conv2d\_6    & $1.0$   & $0.888$ & $0.258$ \\
LeakyReLU\_6 & $1.0$   & $0.922$ & $0.258$ \\ \bottomrule
\end{tabular}
\label{table:innersubsetStarGAN}
\end{table}
\begin{table*}[h!]
\centering
\caption{Precision and recall metrics for group-based scanning over different fake content proportions with 200 runs across $D(x)$ PGGAN layers and CelebA-HQ dataset.}

\begin{tabular}{lllllll}
\toprule
\textbf{layers/prop.}  & \textbf{metric}          & \textbf{50\%}                             & \textbf{40\%}                             & \textbf{30\%}                             & \textbf{20\%}                               & \textbf{10\%}                              \\ \midrule

\multirow{ 2}{*}{\textbf{scaleLayers.0.0} } & precision                        & $0.929 \pm 0.035$                            & $0.895 \pm 0.052$                            & $0.832 \pm 0.076$                                        & $0.719 \pm 0.125$                              & $0.405 \pm 0.153$                             \\
                    & recall & $0.855 \pm 0.051$                            & $0.851 \pm  0.058$                            & $0.867 \pm 0.072$                                        & $0.875 \pm 0.071$                              & $0.886 \pm 0.105$                             \\
\multirow{-1}{*}{{\textbf{scaleLayers.1.0}}} & precision & { $0.941 \pm 0.034$} & { $0.909 \pm 0.046$} & { $0.859 \pm 0.066$} & { $0.759 \pm  0.110$} & { $0.483 \pm  0.162$} \\
{ } & recall & { $0.900 \pm 0.045$} & { $0.900 \pm 0.052$} & { $ 0.904 \pm 0.053$} & { $0.912 \pm  0.063$} & { $0.938 \pm 0.071$} \\

\multirow{-1}{*}{\textbf{scaleLayers.1.1}} & precision                         & $0.942 \pm 0.036$                                        & $0.899 \pm 0.045$                        & $0.841 \pm 0.072$                        & $0.727 \pm 0.105$                          & $0.416 \pm 0.148$                        \\ 
& recall                        & $0.885 \pm 0.046$                                        & $0.891 \pm 0.048$                        & $0.891 \pm 0.058$                        & $0.891 \pm 0.058$                          & $0.899 \pm 0.068$                        \\ 
\multirow{-1}{*}{\textbf{groupScaleZero.0}}     & precision                   & $0.942 \pm 0.037$                                        & $0.907 \pm 0.051$                        & $0.853 \pm 0.071$                        & $0.752 \pm 0.108$                          & $0.482 \pm 0.108$                        \\
& recall                       & $0.828 \pm 0.054$                                        & $0.837 \pm 0.059$                        & $0.840 \pm 0.072$                        & $0.857 \pm 0.081$                          & $0.871 \pm 0.121$                        \\
\multirow{-1}{*}{\textbf{groupScaleZero.1}} & precision                        & $0.942 \pm 0.035$                                        & $0.917 \pm 0.048$                        & $0.868 \pm 0.066$                        & $0.777 \pm 0.104$                          & $0.474 \pm 0.162$                        \\
& recall                      & $0.832 \pm 0.055$                                        & $0.836 \pm 0.065$                        & $0.843 \pm 0.062$                        & $0.845 \pm 0.079$                          & $0.875 \pm 0.111$                        \\
\multirow{-1}{*}{\textbf{decisionLayer}}        & precision               & $0.326 \pm 0.065$                                        & $0.245 \pm 0.065$                        & $0.171 \pm 0.061$                        & $0.110 \pm 0.045$                          & $ 0.048 \pm  0.033$                        \\
& recall                     & $ 0.218 \pm 0.072$                                        & $0.213 \pm 0.082$                        & $0.209 \pm 0.089$                        & $0.205 \pm 0.093$                          & $ 0.196 \pm 0.142$                        \\

\bottomrule
\end{tabular}
\label{tab:precisionpggan}
\end{table*}
\subsection{Detection of partially synthesized samples}
Attribute editing is considered as an example for partially synthesized content. 
We compared the performance of SubsetGAN (group-based and individual) and selected existing methods, FakeSpotter \citep{wangfakespotter} and AutoGAN \citep{zhang2019detecting},  in detecting samples (from CelebA-HQ datset) with an edited attribute.

Tables \ref{tab:sota} and ~\ref{tab:generalclass}
show that the proposed group-based SubsetGAN outperformed FakeSpotter and AutoGAN across both GAN types (PGGAN and StarGAN). Particularly, both  FakeSpotter  and AutoGAN struggled to detect partially synthesized contents drawn from StarGAN. 
Comparatively, FakeSpotter performed higher than AutoGAN, achieving AUC values second to the proposed SubsetGAN that
gains its detection power by efficiently identifying a subset of nodes that deviate away from expected behavior.  Other methods may rely on aggregate changes (i.e. a change in average across all node activations).  This ability to identify a group of nodes maintains high detection power when only part of the image is edited.
Furthermore, SubsetGAN (group) shows that the identified anomalous set of nodes persists across multiple edited samples, which underlines it unique ability to identify patterns of anomalous activations across a group of images.

We also evaluated SubsetGAN across different layers of StarGAN, with more edited attributes and different ratios of synthesized samples in a group. Both individual and group-based scanning resulted in impressive detection performance, particularly in the middle layers of StarGAN as shown in Table~\ref{table:innersubsetStarGAN}. Moreover, SubsetGAN achieved impressive detection performance  when the ratio of synthesized samples is $50\%$. Generally,  SubsetGAN exhibited higher detection of samples even when only  a few attributes are edited.
Similar experiments were performed for an other validation task, i.e. image translation. Specifically, we tested SubsetGAN with CycleGAN dataset \emph{horse2zebra} \citep{CycleGAN2017}, yielding $0.973$ AUC from individual scan, $0.997$ from group-based scanning with $P=0.990$ and $R=0.996$.

\subsection{Detection of completely synthesized samples}
In addition to our validation on detection of partially generated samples, we also evaluated SubsetGAN in detecting completely synthesized samples. 
We used synthesized samples from  PGGAN, real samples from CelebA-HQ dataset and group-based SubsetGAN was applied to detect those synthesized samples. The detection performance across different layers of PGGAN and for different ratios of synthesized samples are shown  in Table~\ref{tab:precisionpggan}. We started with a 50\% proportion because it is a standard procedure to use {\em minibatch discrimination} that looks at multiple examples (real and generated) rather than in isolation,  as this helps avoid a collapse of the generator~\citep{salimans2016improved}. 

Results showed that the intermediate layers had more discriminative activations compared to the other layers, consistent with partially synthesized-content detection using StarGAN (see Table~\ref{table:innersubsetStarGAN}). Compared to other synthesized samples ratios, $50\%$ achieved superior trade-off between precision and recall, as expected, given the more balanced positive and negative samples in the group. Smaller ratios, e.g., $10\%$, exhibited higher recall values due to the lower likelihood of samples being predicted as false negatives. This is validated by the low precision values achieved by smaller ratios due to the higher likelihood of samples being predicted as false positives.     

Figure \ref{fig:rocpgganceleba} shows the ROC curves with their corresponding AUROC across different layers of the discriminator for CelebA-HQ real samples and completely synthesized samples from PGGAN, across several proportions from group-based scan and individual scan.
Table~\ref{tab:precisionpggan} shows precision and recall averaged across 200 runs for group-based scanning over different fake content proportions over the discriminator layers of  PGGAN  and using CelebA-HQ dataset.
We observe across different experiments with DCGAN and PGGAN, that the first layers maintain  high detection power (AUC) when the proportion is higher than 20\% of generated samples, but decays when the amount of generated samples is below or equal to 10\%. In this case, opting for individual scan will yield better detection power.
\subsection{Run-time benchmark}
Scalability is often an issue in most existing anomalous sample detection techniques, particularly, when  the group of samples is large. After all, there are exponentially many subsets with respect to the group size. To this end, SubsetGAN utilises linear-time subset scanning property that helps to scan across samples in linear time via its ranking function.
In Table~\ref{tab:time} we can see the execution time for subset scanning under a convolutional layer ({\em main.2}) with $131,200$ learn-able parameters from PGGAN $D(x)$. 
For the evaluation we performed $200$ runs with the Berk-Jones~\citep{berk-bj-1979} score function. We evaluated the proposed method for single and multiple images (each image is $256 \times 256$ pixels) as input for the network. The tests were performed in a desktop machine (2.9 GHz Quad-Core Intel Core i7, 16 GB 2133 MHz LPDDR3).
\begin{table}[h]
\caption{Benchmark for SubsetGAN. Scan time involves p-value calculation and scanning process for both evaluation samples. Total time measure the complete pipeline from node activation extraction till output metrics recording.}
\centering
\label{tab:time}
\begin{tabular}{@{}ccc@{}}
\toprule
\# images & Scan time (secs) & Total time (secs) \\ \midrule
1                   & $62.0 \pm 0.43$                                   & $67.7\pm 0.49$           \\
10                   & $77.7 \pm 0.41$                                   & $83.8 \pm 0.82$          \\
100                  & $94.41\pm 0.42$                                   & $100.7\pm 0.94$         \\
1000                & $248.5\pm 1.27$                                  & $262.77 \pm1.34$         \\ \bottomrule
\end{tabular}

\end{table}


\section{Conclusion and Future Work}
We proposed SubsetGAN - a novel method to detect AI-synthesized content via subset scanning. 
Unlike SubsetGAN, existing methods to detect synthesized content often require labelled generated examples, re-training of models and/or augmentation of data. 
Our proposed method works by analysing the activation space of the discriminator component of any given generative network or off-the-shelf fake classifier. SubsetGAN provides both the subset of the input images identified as AI-synthesized and the corresponding nodes in the network that gave rise to the identification of those images. With this approach, we aim to enhance the detection power of current deep learning based fake detectors.

SubsetGAN (individual) can operate on a per-image basis and shows strong detection power without requiring multiple images in a test set.  This is because our method does not require aggregate-level changes to the activation space to detect generated content.  However, additional unique insights are gained with the ability to identify anomaolaous nodes across a \emph{group} of images. We validated SubsetGAN across different generative networks (e.g., PGGAN, StarGAN, CycleGAN and DCGAN) and types of generation: \textit{partial} (e.g., attribute editing) and \textit{complete}. Further, we evaluated the case where no information of the generative source is provided, with general fake classifiers, that are trained with samples from multiple GANs and generation types. We compared the detection capability of  SubsetGAN with existing methods FakeSpotter and AutoGAN. The results showed that SubsetGAN outperformed those existing methods consistently across different validation scenarios, and also drives towards interpretability of the detection process. 
Future work includes utilising the characteristics of the anomalous nodes in improving the generative component. 
SubsetGAN also sets the foundation to transition from robust detection of generated content to explainability and retraining of generative models.

\bibliographystyle{ACM-Reference-Format}
\bibliography{main}

\renewcommand{\thesection}{A.\arabic{section}}
\appendix
\section{Supplementary Material}
\subsection{Additional Experiments on baseline DCGAN and MNIST}
\begin{figure}[h!]
    \centering
    \includegraphics[width=0.45\textwidth]{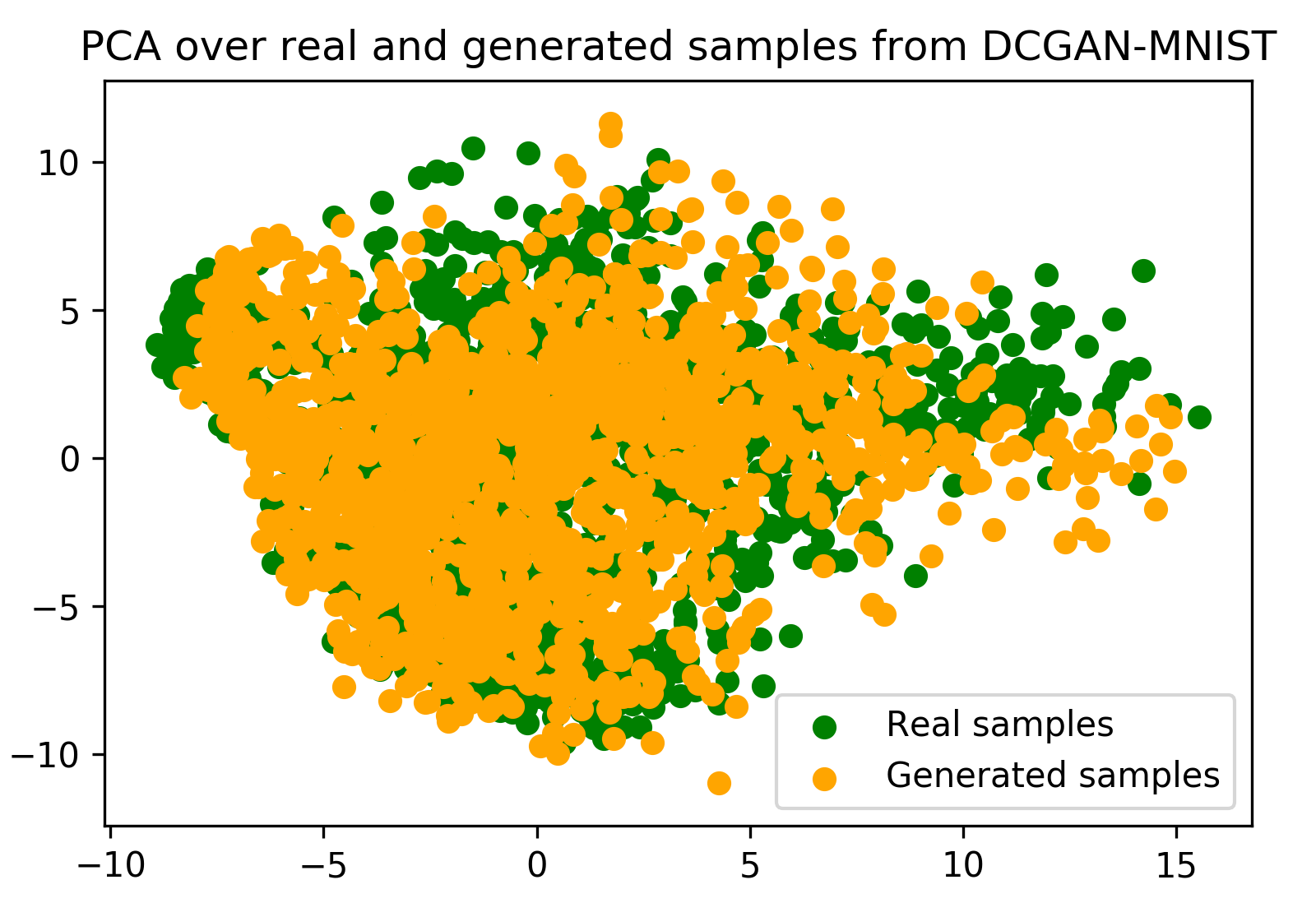}
    \caption{PCA over clean and generated samples from DCGAN-MNIST. To verify that the distributions overlap.}
    \label{fig:pca}
\end{figure}

The architecture of the scanned $D(x)$ from DCGAN can be found in Table~\ref{tab:structure}.
For a qualitative inspection of the samples used in the experiments, we can observe 
a PCA overlapping both generated and real samples can be seen in Figure~\ref{fig:pca}.


\begin{figure}[h]
    \centering
   \includegraphics[width=0.5\textwidth]{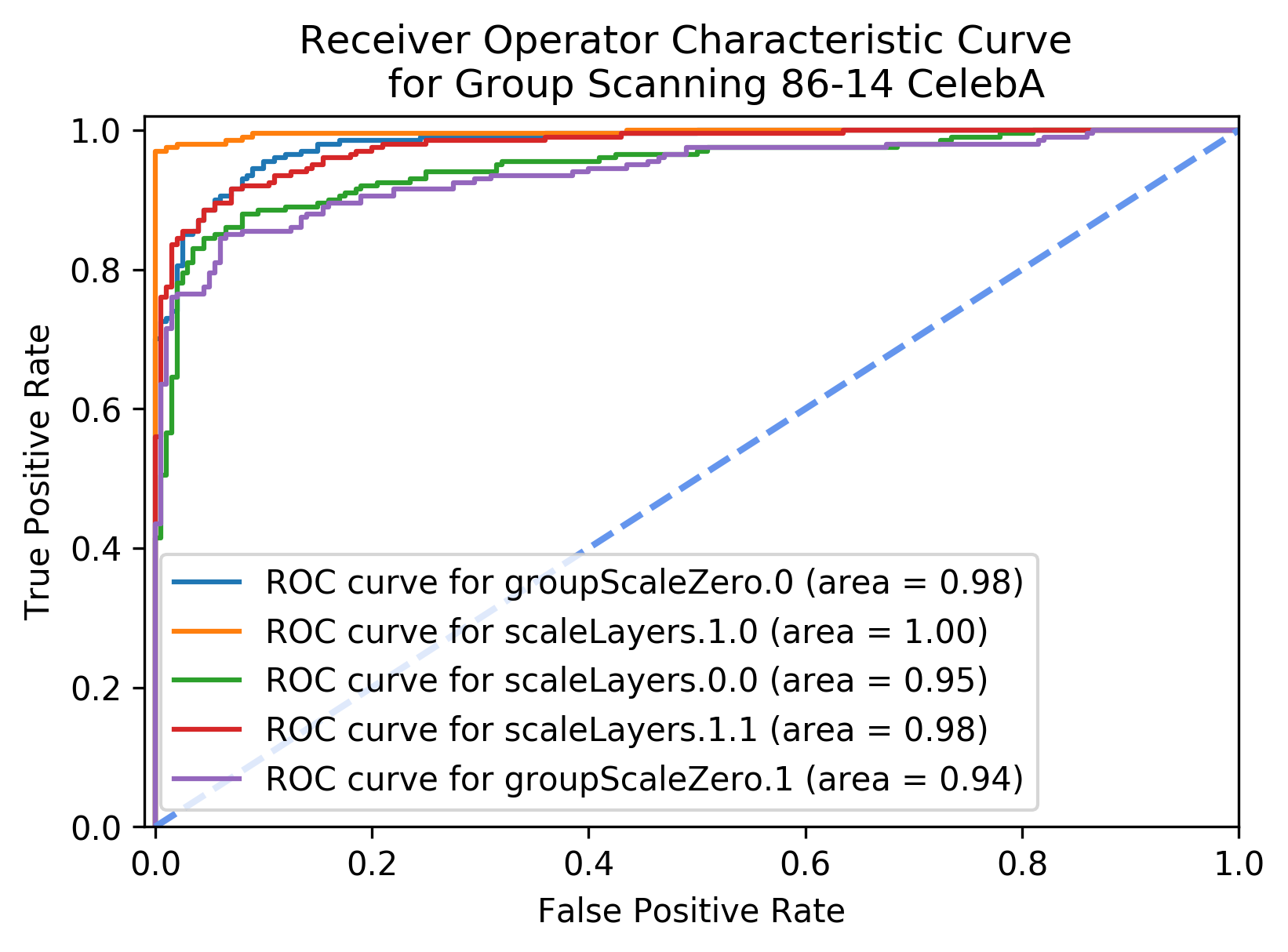}\includegraphics[width=0.5\textwidth]{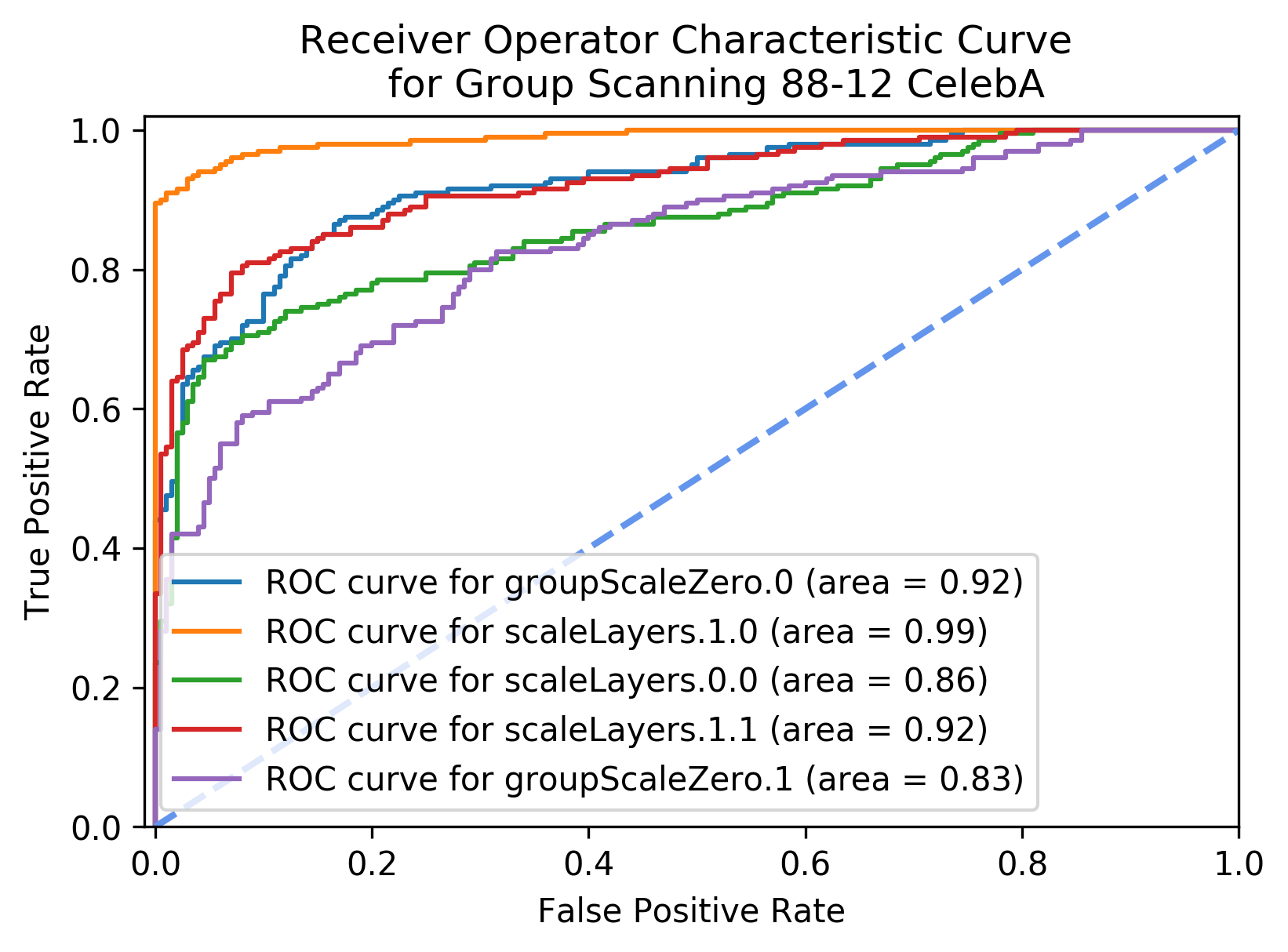}

    \caption{ROC curves for group-based scanning over multiple proportions till 90-10 proportion for PGGAN with CelebA-HQ.}
    \label{fig:rocpggancelebasup}
\end{figure}
\begin{table}[h]
\centering
\caption{$D(x)$ layers and parameters description from DCGAN}
\begin{tabular}{llp{4cm}}
\toprule
Name   & type        & parameters                                   \\ \midrule
main.0 & Conv2d      & kernel\_size=$4\times 4$, stride=$2\times 2$ \\
main.1 & LeakyReLU   & negative\_slope=0.2                          \\
main.2 & Conv2d      & kernel\_size=$4\times 4$, stride=$2\times 2$ \\
main.3 & BatchNorm2d & eps=1e-05, momentum=0.1, affine=True         \\
main.4 & LeakyReLU   & negative\_slope=0.2                          \\
main.5 & Conv2d      & kernel\_size=$4\times 4$, stride=$2\times 2$ \\
main.6 & BatchNorm2d & eps=1e-05, momentum=0.1, affine=True         \\
main.7 & LeakyReLU   & negative\_slope=0.2                          \\
main.8 & Conv2d      & kernel\_size=$4\times 4$, stride=$2\times 2$ \\
main.9 & Sigmoid     & -                                            \\ \bottomrule
\end{tabular}

\label{tab:structure}
\end{table}
\subsection{Additional experiments on completely synthesized samples detection}
In Figure~\ref{fig:rocpggancelebasup} we can observe a more detail behaviour for completely synthesized samples across smaller proportions.

\subsection{Additional experiments on partially synthesized samples detection}
Regarding image translation with \emph{horse2zebra} dataset ~\citep{CycleGAN2017} we report Precision and Recall across layers of CycleGAN $D(x)$ in Table~\ref{table:innersubsetCyclegan}.
\begin{table}[h!]
\centering
\caption{{\bf Precision and Recall} for group-based subset scanning over all layers of CycleGAN $D(x)$ for image translation with \emph{horse2zebra} dataset~\citep{CycleGAN2017}.}
\begin{tabular*}{0.47\textwidth}{lcc}
\toprule
 \multicolumn{1}{c}{Layers} &
      \multicolumn{2}{c}{Group-based Subset scan for $D(x)$}\\
      &
      \multicolumn{2}{c}{20\%}\\
      \midrule
        & {P} & {R}\\
\midrule
Conv2d\_1                                    & $0.959 \pm 0.022$                                             & $0.998\pm 0.007$ \\
LeakyReLU\_1                          & $0.961 \pm 0.023$                                             & $0.997 \pm 0.010$ \\
Conv2d\_2  & $0.990 \pm 0.006$                                             & $0.997 \pm 0.010$  \\
InstanceNorm2d\_1   & $0.860 \pm 0.041$                                             & $ 0.983 \pm 0.025$ \\
LeakyReLU\_2                           & $0.849 \pm 0.036$    & $0.978 \pm 0.026$                                            \\
Conv2d\_3  & $0.990 \pm 0.005$                                             & $0.958 \pm 0.036$ \\
InstanceNorm2d\_2   & $ 0.948 \pm 0.026$                                             & $0.998 \pm 0.007$\\
LeakyReLU\_3                           & $0.953 \pm 0.027$                                            &  $0.997 \pm 0.010$ \\
Conv2d\_5  & $  0.997\pm 0.011$                                             & $ 0.684 \pm 0.099$ \\
InstanceNorm2d\_3   & $ 0.804\pm 0.080$                                             & $0.882 \pm 0.055$\\
LeakyReLU\_4                           & $0.784 \pm 0.068$                                             & $ 0.882\pm 0.058$ \\
\bottomrule
\end{tabular*}
\label{table:innersubsetCyclegan}
\end{table}

\end{document}